\documentclass[10pt,twocolumn,letterpaper]{article}

\usepackage[pagenumbers]{cvpr} 

\usepackage{graphicx}
\usepackage{amsmath}
\usepackage{amsthm}
\usepackage{amssymb}
\usepackage{mathtools}
\usepackage{booktabs}
\usepackage{bm}
\usepackage{multirow}
\usepackage{float}

\usepackage[pagebackref,breaklinks,colorlinks]{hyperref}

\usepackage[capitalize]{cleveref}
\crefname{section}{Sec.}{Secs.}
\Crefname{section}{Section}{Sections}
\Crefname{table}{Table}{Tables}
\crefname{table}{Tab.}{Tabs.}

\begin{document}

\title{EVIMO2: An Event Camera Dataset for Motion Segmentation, Optical Flow, Structure from Motion, and Visual Inertial Odometry in Indoor Scenes with Monocular or Stereo Algorithms}


\author{Levi Burner\thanks{Levi Burner is the corresponding author. The support of the NSF under awards DGE-1632976 and OISE 2020624 is gratefully acknowledged.} , Anton Mitrokhin,  Cornelia Ferm\"{u}ller, Yiannis Aloimonos\\
Perception and Robotics Group, University of Maryland Institute for Advanced Computer Studies\\
University of Maryland, College Park\\
}

\maketitle

\begin{abstract}
A new event camera dataset, EVIMO2, is introduced that improves on the popular EVIMO dataset by providing more data, from better cameras, in more complex scenarios. As with its predecessor, EVIMO2 provides labels in the form of per-pixel ground truth depth and segmentation as well as camera and object poses. All sequences use data from physical cameras and many sequences feature multiple independently moving objects. Typically, such labeled data is unavailable in physical event camera datasets. Thus, EVIMO2 will serve as a challenging benchmark for existing algorithms and rich training set for the development of new algorithms. In particular, EVIMO2 is suited for supporting research in motion and object segmentation, optical flow, structure from motion, and visual (inertial) odometry in both monocular or stereo configurations.

EVIMO2 consists of 41 minutes of data from three 640$\times$480 event cameras, one 2080$\times$1552 classical color camera, inertial measurements from two six axis inertial measurement units, and millimeter accurate object poses from a Vicon motion capture system. The dataset's 173 sequences are arranged into three categories. 3.75 minutes of independently moving household objects, 22.55 minutes of static scenes, and 14.85 minutes of basic motions in shallow scenes. Some sequences were recorded in low-light conditions where conventional cameras fail. Depth and segmentation are provided at 60 Hz for the event cameras and 30 Hz for the classical camera. The masks can be regenerated using open-source code up to rates as high as 200 Hz.

This technical report briefly describes EVIMO2. The full documentation is available online\footnote{\scriptsize{\url{https://better-flow.github.io/evimo}}}. Videos of individual sequences can be sampled on the download page\footnote{\scriptsize{\url{https://better-flow.github.io/evimo/download_evimo_2.html}}}.

\end{abstract}

\begin{figure}
  \centering
  \includegraphics[width=1.0\linewidth]{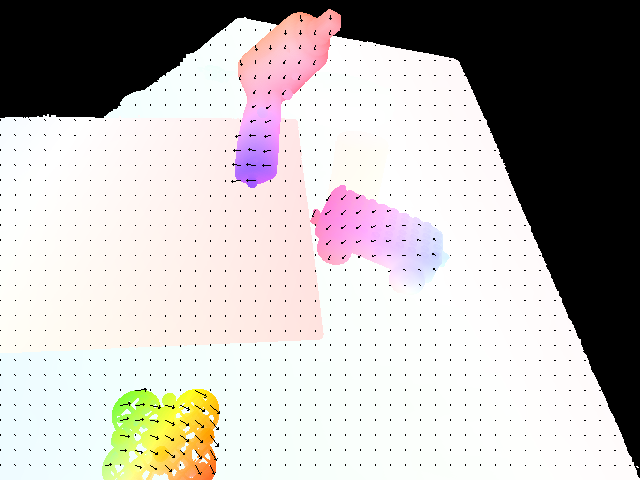}
   \caption{EVIMO2's data can be used to generate high quality ground truth optical flow. Due to Vicon's accurate tracking of independently moving objects at 200 Hz, the flow can be estimated using finite difference over small temporal intervals (10 milliseconds in the figure). The flow is dense as shown in the color field. Sparse arrows indicate magnitude.}
   \label{fig:flowexample}
\end{figure}

\section{Introduction}
\begin{figure*}
  \centering
  \begin{subfigure}{0.44\linewidth}
    \includegraphics[width=1.0\linewidth]{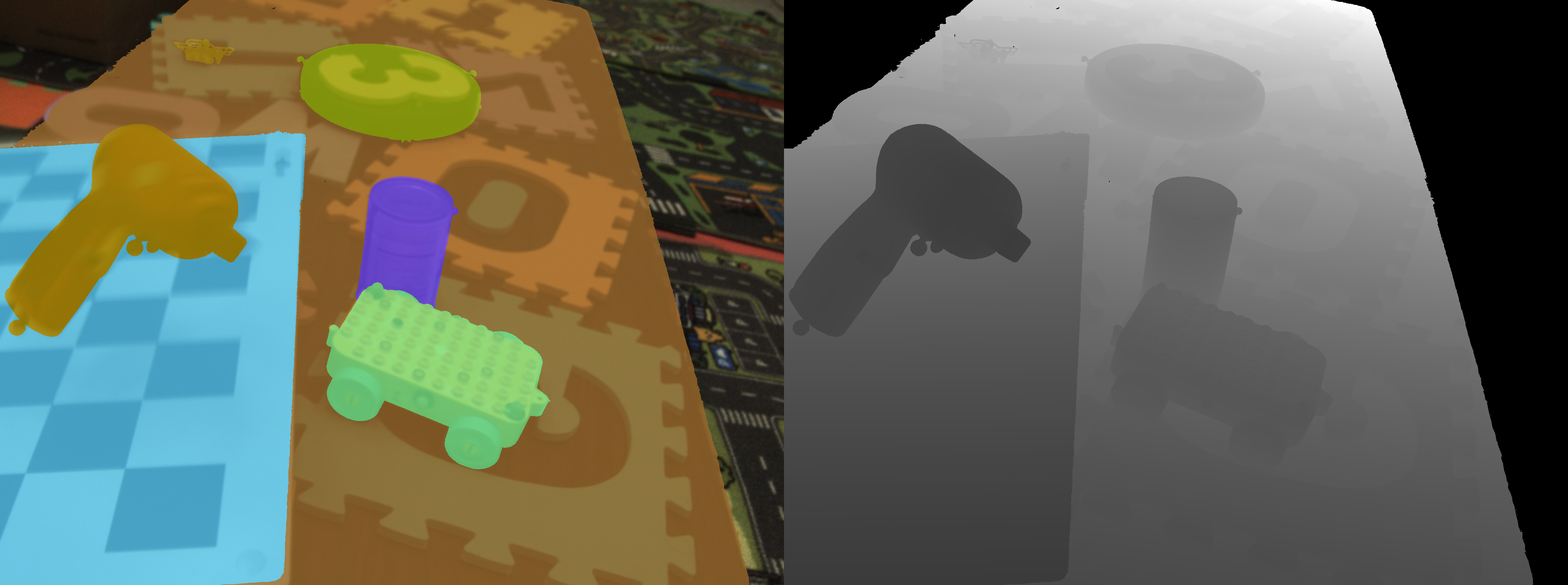}
    \caption{\texttt{Flea3 classical}}
  \end{subfigure}
  \hfill
  \begin{subfigure}{0.44\linewidth}
    \includegraphics[width=1.0\linewidth]{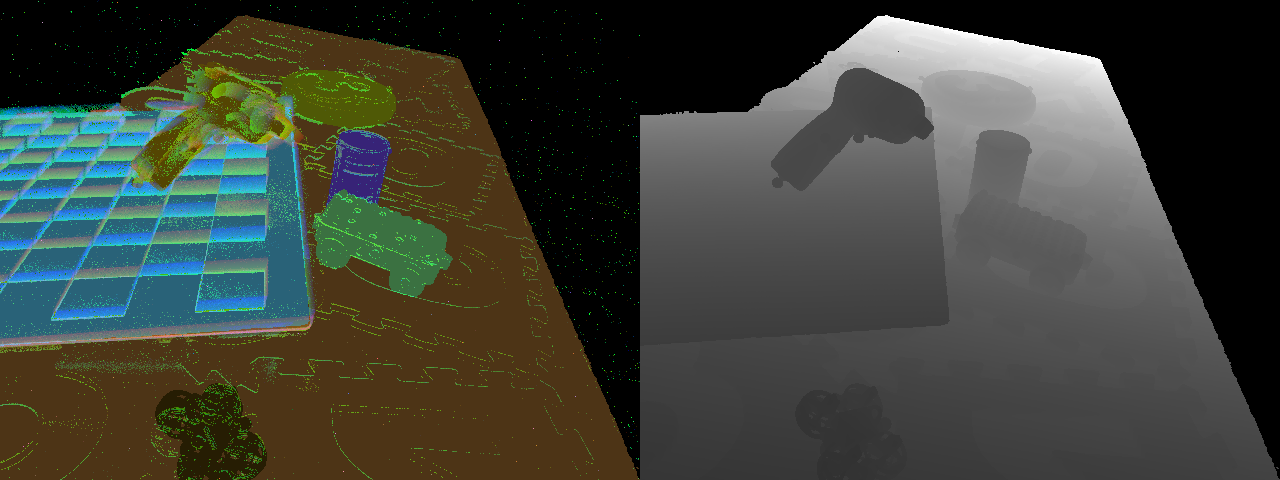}
    \caption{\texttt{Samsung DVS Gen 3}}
  \end{subfigure}
  \hfill
  \begin{subfigure}{0.44\linewidth}
    \includegraphics[width=1.0\linewidth]{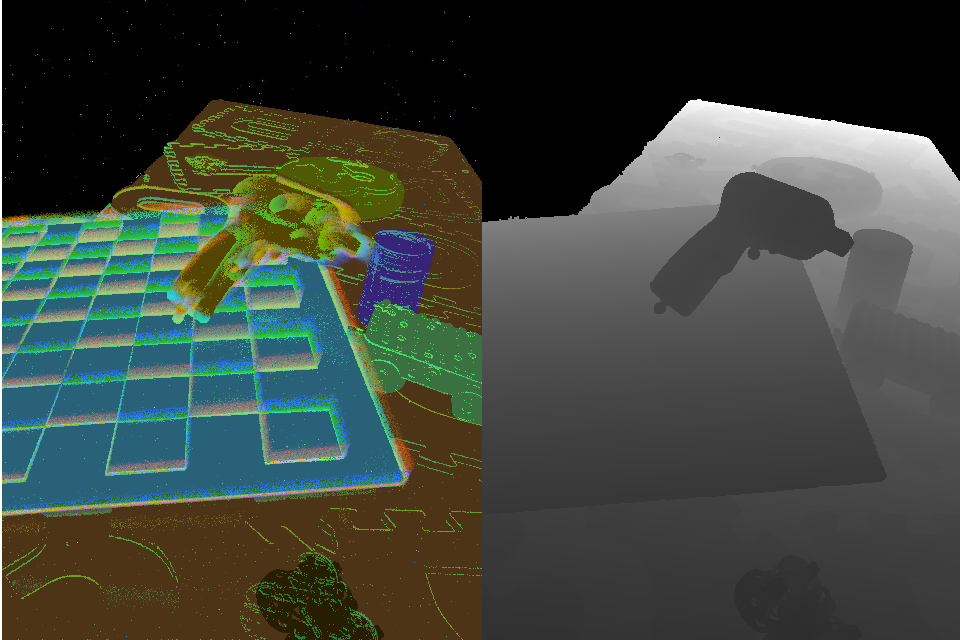}
    \caption{\texttt{Left Prophessee Gen3}}
  \end{subfigure}
  \hfill
  \begin{subfigure}{0.44\linewidth}
    \includegraphics[width=1.0\linewidth]{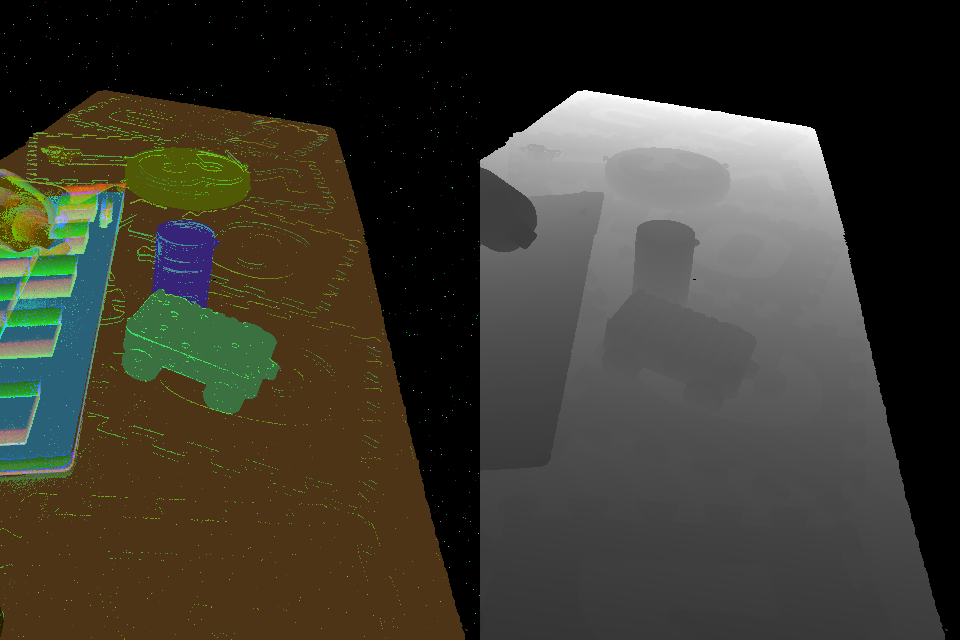}
    \caption{\texttt{Right Prophesee Gen 3}}
  \end{subfigure}
  \caption{In each subframe, the left side visualizes the raw data (events or brightness) with the segmentation mask overlayed as a color. Average event time in a pixel is color coded. The right side of each frame shows the ground truth depth. There are black pixels where depth or segmentation are unavailable.} 
\label{fig:banner}
\end{figure*}

EVIMO2 is an event camera dataset featuring per-pixel ground truth depth, segmentation, and object pose for the footage from three event cameras and one classical color camera. It generally improves on its predecessor EVIMO \cite{Mitrokhin2019EVIMO} by using better cameras, featuring longer recordings, providing different categories of recordings, and using an improved data format. The recordings feature indoor, dynamic scenes with multiple independently moving objects. In contrast to existing simulated \cite{Kaiser2016, mueggler2017, InteriorNet18, pmlrv87rebecq18a, Gehrig_2020_CVPR, Hu2021-v2e-cvpr-workshop-eventvision2021, Nehvi2021, damien2021, airsim2017fsr, GehrigEventSegRAL2021, sanket2020evdodgenet, parameshwara20210} and physical \cite{delbruck2008frame, hu2016dvsbenchmark, moeys2016steering, andreopoulos2018cvprlowpower, zhu2018mvsec, Zhou18eccv, Zhou21tro, Gehrig2021DSEC, bodo2016evaluation, zhu2018mvsec, Almatrafi2020distancesurface, liu2022edflow, Weikersdorfer2014icra, Barranco2016dataset, mueggler2017, binas2017ddd17, zhu2018mvsec, Bryner19icra, Gallego18pami, Delmerico19icra, lee2019vivid, lee2022vivid, hu2020ddd20, Gomez2021griffin, klenk2021tumvie, perot2020learning} event camera datasets, EVIMO2 focuses on close range indoor situations with fast moving objects and provides per-pixel ground truth depth and segmentation as well as poses for the physical cameras and all objects.

While pixel accurate semantic segmentation is sometimes available in simulated event camera datasets \cite{InteriorNet18, airsim2017fsr, GehrigEventSegRAL2021, sanket2020evdodgenet, parameshwara20210} it is not usually available for physical event camera datasets with the exception being EVIMO2's direct predecessor \cite{Mitrokhin2019EVIMO}. On the other hand, while depth and semantic bounding boxes are sometimes present in physical event camera datasets from LIDAR, structured light sensors, and hand labeling \cite{mitrokhin2018event,andreopoulos2018cvprlowpower, zhu2018mvsec, Gehrig2021DSEC, lee2019vivid, lee2022vivid, perot2020learning}, EVIMO2's depth and segmentation comes from 3D scans of household objects combined with Vicon pose measurements. This results in more detailed ground truth depth maps and optical flow fields than could previously be achieved, even in datasets designed for optical flow estimation \cite{bodo2016evaluation, Almatrafi2020distancesurface, Barranco2016dataset, mueggler2017}.

Due to the availability of ground truth pose of the camera, EVIMO2 is suitable for indoor Visual Inertial Odometry (VIO) and Simultaneous Localization and Mapping (SLAM) research. It differs from existing VIO and SLAM event camera datasets by focusing on close-up indoor scenes containing moving household objects. In contrast, existing datasets focus on indoor scenes without moving objects \cite{Barranco2016dataset, Weikersdorfer2014icra, mueggler2017, Bryner19icra, Gallego18pami, klenk2021tumvie}, outdoor or driving/flying scenes \cite{binas2017ddd17, zhu2018mvsec, Delmerico19icra, hu2020ddd20, Gomez2021griffin, klenk2021tumvie}, or extreme environments \cite{lee2019vivid, lee2022vivid, mitrokhin2018event}. Further, EVIMO2 provides poses for the camera and all objects which opens up exciting opportunities for object pose estimation.

EVIMO2's ground truth depth and segmentation masks are obtained using the same methods as its predecessor EVIMO \cite{Mitrokhin2019EVIMO}. The 3D scans of the objects are projected into the camera's field-of-view using poses measured by a Vicon motion capture system. 60 Hz ground truth is available for event cameras and 30 Hz ground truth for the classical camera. Higher framerate ground truth can be generated using the released open-source tools. Due to the detailed 3D object scans and the high quality tracking of the Vicon system, exceptional ground-truth optical flow can be generated for scenes with fast moving objects undergoing complex motions as shown in Figure \ref{fig:flowexample} and \ref{fig:flow2}.

\section{Methods}
EVIMO2 features sensor data from three event cameras, one classical camera, two six-axis inertial measurement units, and poses from a Vicon motion capture system. Two of the event cameras are Prophesee Gen3 VGA 640$\times$480 sensors with a 71 degree diagonal field-of-view \footnote{\scriptsize{\url{https://docs.prophesee.ai/stable/hw/evk/gen3.html}}}. They are arranged in a binocular stereo configuration with a  baseline of approximately 22 cm. The third event camera is a Samsung DVS Gen3 with 640$\times$480 resolution and 75 degree diagonal FOV \footnote{\scriptsize{\url{https://rpg.ifi.uzh.ch/docs/CVPR19workshop/CVPRW19_Eric_Ryu_Samsung.pdf}}}. The classical camera is a Flea 3 camera featuring the Sony IMX036 sensor which recorded at 2080$\times$1552 and 30 Hz while fitted with a lens that achieved a 64 degree diagonal FOV \footnote{\scriptsize{\url{https://www.flir.com/products/flea3-usb3/}}}. The inertial measurements are supplied by the Prophesee camera's internal MPU9250 IMUs which are sampled at 1 kHz \footnote{\scriptsize{\url{https://support.prophesee.ai/portal/en/kb/articles/imu-data-format-in-metavision-software}}}.

\subsection{Objects}

The dataset features over 20 real world objects including a large table that forms the main backdrop of many sequences as illustrated by Figure \ref{fig:banner}. The other objects are  small household items such as toy blocks, remote control cars, and small drones. Silver motion capture markers are attached to each object to enable tracking by the Vicon motion capture system.

The 3D Models for the objects were obtained using an Artec Space Spider scanner\footnote{\scriptsize{\url{https://www.artec3d.com/portable-3d-scanners/artec-spider}}}. This results in high quality meshes, which were then edited using the open-source Meshlab software to remove the Vicon markers \cite{meshlab2008}.

\subsection{Ground Truth}
Ground truth depth and segmentation masks are calculated using Vicon pose estimates and 3D scans of small objects and a large table. This makes accurate ground truth depth available for the objects and the table, which typically cover most of the field of view as exemplified by Figure \ref{fig:banner}.

Due to the small size of the objects used, and the sensitivity of the Vicon motion capture system to occlusion of  object's markers, there are occasional temporal gaps in ground truth availability. As a result, users must take care to respect that the ground truth data is not evenly spaced in time.

\subsection{Sensor Availability}
Not all sensors are available in all sequences. A complete camera availability matrix is available on the EVIMO2 downloads page. The left Prophesee camera is available in 155 sequences. The right Prophesee camera is available in 124 sequences. The Samsung camera is available in 154 sequences. The Flea3 camera is available in 139 sequences (the Flea3 data is excluded from several of the low-light sequences because it did not produce usable data).

\subsection{Sequence Categories}

The sequences are organized in three distinct categories. ``Independently Moving Objects'' (IMO), ``Structure from Motion'' (SfM), and ``Simple Motion in Planar Scenes'' (Sanity). All categories have a smaller set of sequences designated ``low light''. In low light sequences the ambient light levels in the recording room were greatly reduced which makes the events much ``noisier''. Video previews of a prototypical member of each category are available on EVIMO2's download page.

For IMO and SfM sequences, sequences have been split into distinct ``train'' and ``test'' sets of approximately equal length. The sanity sequences are split into ``checkerboard'', ``depth variation'', ``sliding'', and ``tabletop'' subsets where basic translations and rotations are performed in almost entirely planar scenes.

Table \ref{tab:summarystats} summarizes the number and duration of the sequences in each category.

\subsection{Optical Flow}

\begin{figure}
  \centering
  \includegraphics[width=1.0\linewidth]{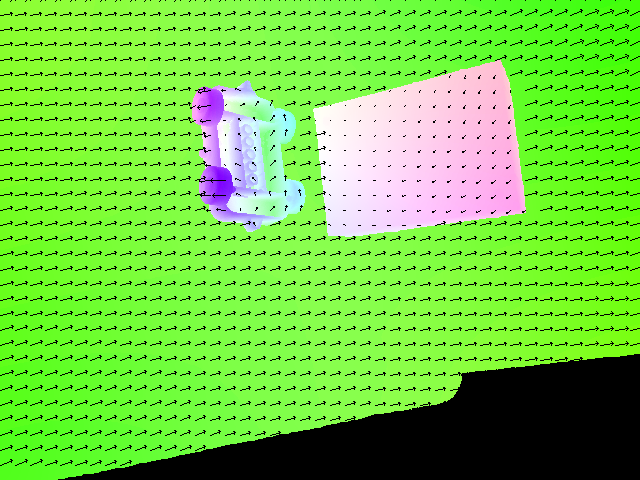}
   \caption{EVIMO2's 3D models allow fine details in the optical flow fields to be captured such as the differences in velocity of the features on a toy car as shown above.}
   \label{fig:flow2}
\end{figure}

Due to the availability of ground truth depth and per object pose, EVIMO2 is particularly suited for evaluation of optical flow methods. An example of the quality of the ground truth optical flow is shown in Figure \ref{fig:flowexample}. There, the ground truth flow is approximated by  finite-differencing object positions within a time interval  of 10 milliseconds. This small time interval is made possible due to the Vicon's 200 Hz update rate. The result is accurate ground-truth optical flow even when multiple objects are flipping through the air in different directions as illustrated by Figure \ref{fig:flowexample}. Further, the high-quality 3D object models allow fine details to be captured in the flow field as shown in Figure \ref{fig:flow2}.

\begin{table}
  \centering
  \begin{tabular}{@{}lcccccccccc@{}}
    \toprule
 \texttt{Category} & \multicolumn{2}{c}{\texttt{Sequences}} & \multicolumn{2}{c}{\texttt{Minutes}}  \\
 & \texttt{Train} & \texttt{Test} & \texttt{Train} & \texttt{Test} \\
    \midrule
    IMO & 21  &  8 &  2.74 & 0.51\\
    IMO LL                         &  3  &  2 &  0.36 & 0.14\\
    SfM           & 52  & 10 & 16.17 & 2.45\\
    SfM LL                         &  5  &  6 &  2.56 & 1.37\\
    Sanity  & N/A & 31 &   N/A & 6.96\\
    Sanity LL                      & N/A & 35 &   N/A & 7.89\\
    \midrule
    Totals & 81 & 92 & 21.83 & 19.32\\
    \bottomrule
  \end{tabular}
  \caption{Number of sequences and duration of sequences for each sequence category in EVIMO2. Abbreviations: Independent Moving Objects (IMO), Structure from Motion (SfM), Simple Motion in Planar Scenes (Sanity), Low Light (LL).}
  \label{tab:summarystats}
\end{table}

\section{Conclusion}
EVIMO2 is an event camera dataset featuring three event cameras, one classical camera, two IMUs, per-pixel ground truth depth, segmentation, and object poses. Typically, ground segmentation and object poses are only available in simulated datasets. Further, EVIMO2 contains numerous sequences with small, independent, fast moving objects, which present a challenge for optical flow, visual odometry, and SLAM methods. Thus, EVIMO2 provides a challenging benchmark for existing methods and rich set of sequences for training new ones.

\section{Acknowledgements}
The authors thank Tobi Delbruck for assisting in testing and debugging the ground truth optical flow generation tool during the writing of \cite{liu2022edflow}. The authors also thank Suteerth Vishnu for assisting with collection and curation of the data.



{\small
\bibliographystyle{ieee_fullname}
\bibliography{ms_evimo2_arxiv_v1}
}

\end{document}